# Reinforcement Learning: A Survey

**Leslie Pack Kaelbling**                                    LPK@CS.BROWN.EDU
**Michael L. Littman**                                      MLITTMAN@CS.BROWN.EDU
*Computer Science Department, Box 1910, Brown University*
*Providence, RI 02912-1910 USA*

**Andrew W. Moore**                                          AWM@CS.CMU.EDU
*Smith Hall 221, Carnegie Mellon University, 5000 Forbes Avenue*
*Pittsburgh, PA 15213 USA*

## Abstract

This paper surveys the field of reinforcement learning from a computer-science per-
spective. It is written to be accessible to researchers familiar with machine learning. Both
the historical basis of the field and a broad selection of current work are summarized.
Reinforcement learning is the problem faced by an agent that learns behavior through
trial-and-error interactions with a dynamic environment. The work described here has a
resemblance to work in psychology, but differs considerably in the details and in the use
of the word "reinforcement." The paper discusses central issues of reinforcement learning,
including trading off exploration and exploitation, establishing the foundations of the field
via Markov decision theory, learning from delayed reinforcement, constructing empirical
models to accelerate learning, making use of generalization and hierarchy, and coping with
hidden state. It concludes with a survey of some implemented systems and an assessment
of the practical utility of current methods for reinforcement learning.

## 1. Introduction

Reinforcement learning dates back to the early days of cybernetics and work in statistics,
psychology, neuroscience, and computer science. In the last five to ten years, it has attracted
rapidly increasing interest in the machine learning and artificial intelligence communities.
Its promise is beguiling—a way of programming agents by reward and punishment without
needing to specify *how* the task is to be achieved. But there are formidable computational
obstacles to fulfilling the promise.

This paper surveys the historical basis of reinforcement learning and some of the current
work from a computer science perspective. We give a high-level overview of the field and a
taste of some specific approaches. It is, of course, impossible to mention all of the important
work in the field; this should not be taken to be an exhaustive account.

Reinforcement learning is the problem faced by an agent that must learn behavior
through trial-and-error interactions with a dynamic environment. The work described here
has a strong family resemblance to eponymous work in psychology, but differs considerably
in the details and in the use of the word "reinforcement." It is appropriately thought of as
a class of problems, rather than as a set of techniques.

There are two main strategies for solving reinforcement-learning problems. The first is to
search in the space of behaviors in order to find one that performs well in the environment.
This approach has been taken by work in genetic algorithms and genetic programming,

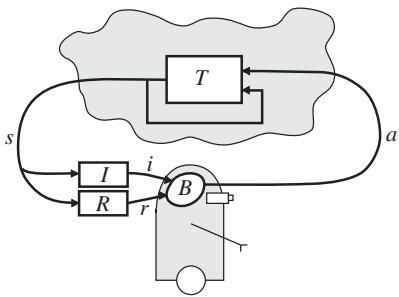

Figure 1: The standard reinforcement-learning model.

as well as some more novel search techniques (Schmidhuber, 1996). The second is to use statistical techniques and dynamic programming methods to estimate the utility of taking actions in states of the world. This paper is devoted almost entirely to the second set of techniques because they take advantage of the special structure of reinforcement-learning problems that is not available in optimization problems in general. It is not yet clear which set of approaches is best in which circumstances.

The rest of this section is devoted to establishing notation and describing the basic reinforcement-learning model. Section 2 explains the trade-off between exploration and exploitation and presents some solutions to the most basic case of reinforcement-learning problems, in which we want to maximize the immediate reward. Section 3 considers the more general problem in which rewards can be delayed in time from the actions that were crucial to gaining them. Section 4 considers some classic model-free algorithms for reinforcement learning from delayed reward: adaptive heuristic critic, $TD(\lambda)$ and Q-learning. Section 5 demonstrates a continuum of algorithms that are sensitive to the amount of computation an agent can perform between actual steps of action in the environment. Generalization—the cornerstone of mainstream machine learning research—has the potential of considerably aiding reinforcement learning, as described in Section 6. Section 7 considers the problems that arise when the agent does not have complete perceptual access to the state of the environment. Section 8 catalogs some of reinforcement learning's successful applications. Finally, Section 9 concludes with some speculations about important open problems and the future of reinforcement learning.

## 1.1 Reinforcement-Learning Model

In the standard reinforcement-learning model, an agent is connected to its environment via perception and action, as depicted in Figure 1. On each step of interaction the agent receives as input, $i$, some indication of the current state, $s$, of the environment; the agent then chooses an action, $a$, to generate as output. The action changes the state of the environment, and the value of this state transition is communicated to the agent through a scalar *reinforcement signal*, $r$. The agent's behavior, $B$, should choose actions that tend to increase the long-run sum of values of the reinforcement signal. It can learn to do this over time by systematic trial and error, guided by a wide variety of algorithms that are the subject of later sections of this paper.

Formally, the model consists of

- a discrete set of environment states, $\mathcal{S}$;

- a discrete set of agent actions, $\mathcal{A}$; and

- a set of scalar reinforcement signals; typically $\{0, 1\}$, or the real numbers.

The figure also includes an input function $I$, which determines how the agent views the environment state; we will assume that it is the identity function (that is, the agent perceives the exact state of the environment) until we consider partial observability in Section 7.

An intuitive way to understand the relation between the agent and its environment is with the following example dialogue.

| | |
|---|---|
| **Environment:** | You are in state 65. You have 4 possible actions. |
| **Agent:** | I'll take action 2. |
| **Environment:** | You received a reinforcement of 7 units. You are now in state 15. You have 2 possible actions. |
| **Agent:** | I'll take action 1. |
| **Environment:** | You received a reinforcement of -4 units. You are now in state 65. You have 4 possible actions. |
| **Agent:** | I'll take action 2. |
| **Environment:** | You received a reinforcement of 5 units. You are now in state 44. You have 5 possible actions. |
| $\vdots$ | $\vdots$ |

The agent's job is to find a policy $\pi$, mapping states to actions, that maximizes some long-run measure of reinforcement. We expect, in general, that the environment will be non-deterministic; that is, that taking the same action in the same state on two different occasions may result in different next states and/or different reinforcement values. This happens in our example above: from state 65, applying action 2 produces differing reinforcements and differing states on two occasions. However, we assume the environment is stationary; that is, that the *probabilities* of making state transitions or receiving specific reinforcement signals do not change over time.[1]

Reinforcement learning differs from the more widely studied problem of supervised learning in several ways. The most important difference is that there is no presentation of input/output pairs. Instead, after choosing an action the agent is told the immediate reward and the subsequent state, but *not* told which action would have been in its best long-term interests. It is necessary for the agent to gather useful experience about the possible system states, actions, transitions and rewards actively to act optimally. Another difference from supervised learning is that on-line performance is important: the evaluation of the system is often concurrent with learning.

---

1. This assumption may be disappointing; after all, operation in non-stationary environments is one of the motivations for building learning systems. In fact, many of the algorithms described in later sections are effective in slowly-varying non-stationary environments, but there is very little theoretical analysis in this area.

Some aspects of reinforcement learning are closely related to search and planning issues in artificial intelligence. AI search algorithms generate a satisfactory trajectory through a graph of states. Planning operates in a similar manner, but typically within a construct with more complexity than a graph, in which states are represented by compositions of logical expressions instead of atomic symbols. These AI algorithms are less general than the reinforcement-learning methods, in that they require a predefined model of state transitions, and with a few exceptions assume determinism. On the other hand, reinforcement learning, at least in the kind of discrete cases for which theory has been developed, assumes that the entire state space can be enumerated and stored in memory—an assumption to which conventional search algorithms are not tied.

## 1.2 Models of Optimal Behavior

Before we can start thinking about algorithms for learning to behave optimally, we have to decide what our model of optimality will be. In particular, we have to specify how the agent should take the future into account in the decisions it makes about how to behave now. There are three models that have been the subject of the majority of work in this area.

The *finite-horizon* model is the easiest to think about; at a given moment in time, the agent should optimize its expected reward for the next $h$ steps:

$$E(\sum_{t=0}^{h} r_t) \ ;$$

it need not worry about what will happen after that. In this and subsequent expressions, $r_t$ represents the scalar reward received $t$ steps into the future. This model can be used in two ways. In the first, the agent will have a non-stationary policy; that is, one that changes over time. On its first step it will take what is termed a *h-step optimal action*. This is defined to be the best action available given that it has $h$ steps remaining in which to act and gain reinforcement. On the next step it will take a $(h-1)$-step optimal action, and so on, until it finally takes a 1-step optimal action and terminates. In the second, the agent does *receding-horizon control*, in which it always takes the $h$-step optimal action. The agent always acts according to the same policy, but the value of $h$ limits how far ahead it looks in choosing its actions. The finite-horizon model is not always appropriate. In many cases we may not know the precise length of the agent's life in advance.

The infinite-horizon discounted model takes the long-run reward of the agent into account, but rewards that are received in the future are geometrically discounted according to discount factor $\gamma$, (where $0 \leq \gamma < 1$):

$$E(\sum_{t=0}^{\infty} \gamma^t r_t) \ .$$

We can interpret $\gamma$ in several ways. It can be seen as an interest rate, a probability of living another step, or as a mathematical trick to bound the infinite sum. The model is conceptually similar to receding-horizon control, but the discounted model is more mathematically tractable than the finite-horizon model. This is a dominant reason for the wide attention this model has received.

Another optimality criterion is the *average-reward model*, in which the agent is supposed to take actions that optimize its long-run average reward:

$$\lim_{h \to \infty} E(\frac{1}{h} \sum_{t=0}^{h} r_t) \ \ .$$

Such a policy is referred to as a *gain optimal* policy; it can be seen as the limiting case of the infinite-horizon discounted model as the discount factor approaches 1 (Bertsekas, 1995). One problem with this criterion is that there is no way to distinguish between two policies, one of which gains a large amount of reward in the initial phases and the other of which does not. Reward gained on any initial prefix of the agent's life is overshadowed by the long-run average performance. It is possible to generalize this model so that it takes into account both the long run average and the amount of initial reward than can be gained. In the generalized, *bias optimal* model, a policy is preferred if it maximizes the long-run average and ties are broken by the initial extra reward.

Figure 2 contrasts these models of optimality by providing an environment in which changing the model of optimality changes the optimal policy. In this example, circles represent the states of the environment and arrows are state transitions. There is only a single action choice from every state except the start state, which is in the upper left and marked with an incoming arrow. All rewards are zero except where marked. Under a finite-horizon model with $h = 5$, the three actions yield rewards of $+6.0$, $+0.0$, and $+0.0$, so the first action should be chosen; under an infinite-horizon discounted model with $\gamma = 0.9$, the three choices yield $+16.2$, $+59.0$, and $+58.5$ so the second action should be chosen; and under the average reward model, the third action should be chosen since it leads to an average reward of $+11$. If we change $h$ to 1000 and $\gamma$ to 0.2, then the second action is optimal for the finite-horizon model and the first for the infinite-horizon discounted model; however, the average reward model will always prefer the best long-term average. Since the choice of optimality model and parameters matters so much, it is important to choose it carefully in any application.

The finite-horizon model is appropriate when the agent's lifetime is known; one important aspect of this model is that as the length of the remaining lifetime decreases, the agent's policy may change. A system with a hard deadline would be appropriately modeled this way. The relative usefulness of infinite-horizon discounted and bias-optimal models is still under debate. Bias-optimality has the advantage of not requiring a discount parameter; however, algorithms for finding bias-optimal policies are not yet as well-understood as those for finding optimal infinite-horizon discounted policies.

## 1.3 Measuring Learning Performance

The criteria given in the previous section can be used to assess the policies learned by a given algorithm. We would also like to be able to evaluate the quality of learning itself. There are several incompatible measures in use.

- **Eventual convergence to optimal.** Many algorithms come with a provable guarantee of asymptotic convergence to optimal behavior (Watkins & Dayan, 1992). This is reassuring, but useless in practical terms. An agent that quickly reaches a plateau at 99% of optimality may, in many applications, be preferable to an agent that has a guarantee of eventual optimality but a sluggish early learning rate.

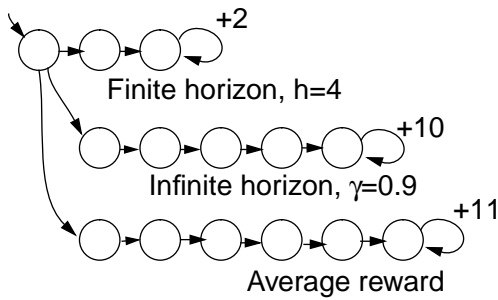

Figure 2: Comparing models of optimality. All unlabeled arrows produce a reward of zero.

- **Speed of convergence to optimality.** Optimality is usually an asymptotic result, and so convergence speed is an ill-defined measure. More practical is the *speed of convergence to near-optimality*. This measure begs the definition of how near to optimality is sufficient. A related measure is *level of performance after a given time*, which similarly requires that someone define the given time.

  It should be noted that here we have another difference between reinforcement learning and conventional supervised learning. In the latter, expected future predictive accuracy or statistical efficiency are the prime concerns. For example, in the well-known PAC framework (Valiant, 1984), there is a learning period during which mistakes do not count, then a performance period during which they do. The framework provides bounds on the necessary length of the learning period in order to have a probabilistic guarantee on the subsequent performance. That is usually an inappropriate view for an agent with a long existence in a complex environment.

  In spite of the mismatch between embedded reinforcement learning and the train/test perspective, Fiechter (1994) provides a PAC analysis for Q-learning (described in Section 4.2) that sheds some light on the connection between the two views.

  Measures related to speed of learning have an additional weakness. An algorithm that merely tries to achieve optimality as fast as possible may incur unnecessarily large penalties during the learning period. A less aggressive strategy taking longer to achieve optimality, but gaining greater total reinforcement during its learning might be preferable.

- **Regret.** A more appropriate measure, then, is the expected decrease in reward gained due to executing the learning algorithm instead of behaving optimally from the very beginning. This measure is known as *regret* (Berry & Fristedt, 1985). It penalizes mistakes wherever they occur during the run. Unfortunately, results concerning the regret of algorithms are quite hard to obtain.

## 1.4 Reinforcement Learning and Adaptive Control

Adaptive control (Burghes & Graham, 1980; Stengel, 1986) is also concerned with algorithms for improving a sequence of decisions from experience. Adaptive control is a much more mature discipline that concerns itself with dynamic systems in which states and actions are vectors and system dynamics are smooth: linear or locally linearizable around a desired trajectory. A very common formulation of cost functions in adaptive control are quadratic penalties on deviation from desired state and action vectors. Most importantly, although the dynamic model of the system is not known in advance, and must be estimated from data, the *structure* of the dynamic model is fixed, leaving model estimation as a parameter estimation problem. These assumptions permit deep, elegant and powerful mathematical analysis, which in turn lead to robust, practical, and widely deployed adaptive control algorithms.

## 2. Exploitation versus Exploration: The Single-State Case

One major difference between reinforcement learning and supervised learning is that a reinforcement-learner must explicitly explore its environment. In order to highlight the problems of exploration, we treat a very simple case in this section. The fundamental issues and approaches described here will, in many cases, transfer to the more complex instances of reinforcement learning discussed later in the paper.

The simplest possible reinforcement-learning problem is known as the $k$-armed bandit problem, which has been the subject of a great deal of study in the statistics and applied mathematics literature (Berry & Fristedt, 1985). The agent is in a room with a collection of $k$ gambling machines (each called a "one-armed bandit" in colloquial English). The agent is permitted a fixed number of pulls, $h$. Any arm may be pulled on each turn. The machines do not require a deposit to play; the only cost is in wasting a pull playing a suboptimal machine. When arm $i$ is pulled, machine $i$ pays off 1 or 0, according to some underlying probability parameter $p_i$, where payoffs are independent events and the $p_i$s are unknown. What should the agent's strategy be?

This problem illustrates the fundamental tradeoff between exploitation and exploration. The agent might believe that a particular arm has a fairly high payoff probability; should it choose that arm all the time, or should it choose another one that it has less information about, but seems to be worse? Answers to these questions depend on how long the agent is expected to play the game; the longer the game lasts, the worse the consequences of prematurely converging on a sub-optimal arm, and the more the agent should explore.

There is a wide variety of solutions to this problem. We will consider a representative selection of them, but for a deeper discussion and a number of important theoretical results, see the book by Berry and Fristedt (1985). We use the term "action" to indicate the agent's choice of arm to pull. This eases the transition into delayed reinforcement models in Section 3. It is very important to note that bandit problems fit our definition of a reinforcement-learning environment with a single state with only self transitions.

Section 2.1 discusses three solutions to the basic one-state bandit problem that have formal correctness results. Although they can be extended to problems with real-valued rewards, they do not apply directly to the general multi-state delayed-reinforcement case.

Section 2.2 presents three techniques that are not formally justified, but that have had wide use in practice, and can be applied (with similar lack of guarantee) to the general case.

## 2.1 Formally Justified Techniques

There is a fairly well-developed formal theory of exploration for very simple problems. Although it is instructive, the methods it provides do not scale well to more complex problems.

### 2.1.1 Dynamic-Programming Approach

If the agent is going to be acting for a total of $h$ steps, it can use basic Bayesian reasoning to solve for an optimal strategy (Berry & Fristedt, 1985). This requires an assumed prior joint distribution for the parameters $\{p_i\}$, the most natural of which is that each $p_i$ is independently uniformly distributed between 0 and 1. We compute a mapping from *belief states* (summaries of the agent's experiences during this run) to actions. Here, a belief state can be represented as a tabulation of action choices and payoffs: $\{n_1, w_1, n_2, w_2, \ldots, n_k, w_k\}$ denotes a state of play in which each arm $i$ has been pulled $n_i$ times with $w_i$ payoffs. We write $V^*(n_1, w_1, \ldots, n_k, w_k)$ as the expected payoff remaining, given that a total of $h$ pulls are available, and we use the remaining pulls optimally.

If $\sum_i n_i = h$, then there are no remaining pulls, and $V^*(n_1, w_1, \ldots, n_k, w_k) = 0$. This is the basis of a recursive definition. If we know the $V^*$ value for all belief states with $t$ pulls remaining, we can compute the $V^*$ value of any belief state with $t + 1$ pulls remaining:

$$
\begin{aligned}
V^*(n_1, w_1, \ldots, n_k, w_k) &= \max_i E \left[ \begin{array}{l} \text{Future payoff if agent takes action } i, \\ \text{then acts optimally for remaining pulls} \end{array} \right] \\
&= \max_i \left( \begin{array}{l} \rho_i V^*(n_1, w_i, \ldots, n_i + 1, w_i + 1, \ldots, n_k, w_k) + \\ (1 - \rho_i) V^*(n_1, w_i, \ldots, n_i + 1, w_i, \ldots, n_k, w_k) \end{array} \right)
\end{aligned}
$$

where $\rho_i$ is the posterior subjective probability of action $i$ paying off given $n_i$, $w_i$ and our prior probability. For the uniform priors, which result in a beta distribution, $\rho_i = (w_i + 1)/(n_i + 2)$.

The expense of filling in the table of $V^*$ values in this way for all attainable belief states is linear in the number of belief states times actions, and thus exponential in the horizon.

### 2.1.2 Gittins Allocation Indices

Gittins gives an "allocation index" method for finding the optimal choice of action at each step in $k$-armed bandit problems (Gittins, 1989). The technique only applies under the discounted expected reward criterion. For each action, consider the number of times it has been chosen, $n$, versus the number of times it has paid off, $w$. For certain discount factors, there are published tables of "index values," $I(n, w)$ for each pair of $n$ and $w$. Look up the index value for each action $i$, $I(n_i, w_i)$. It represents a comparative measure of the combined value of the expected payoff of action $i$ (given its history of payoffs) and the value of the information that we would get by choosing it. Gittins has shown that choosing the action with the largest index value guarantees the optimal balance between exploration and exploitation.

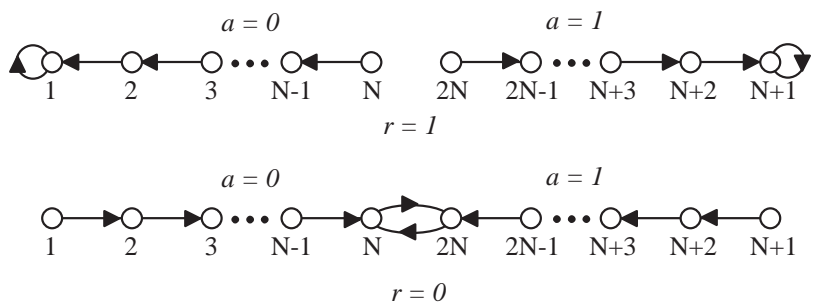

Figure 3: A Tsetlin automaton with $2N$ states. The top row shows the state transitions that are made when the previous action resulted in a reward of 1; the bottom row shows transitions after a reward of 0. In states in the left half of the figure, action 0 is taken; in those on the right, action 1 is taken.

Because of the guarantee of optimal exploration and the simplicity of the technique (given the table of index values), this approach holds a great deal of promise for use in more complex applications. This method proved useful in an application to robotic manipulation with immediate reward (Salganicoff & Ungar, 1995). Unfortunately, no one has yet been able to find an analog of index values for delayed reinforcement problems.

### 2.1.3 LEARNING AUTOMATA

A branch of the theory of adaptive control is devoted to *learning automata*, surveyed by Narendra and Thathachar (1989), which were originally described explicitly as finite state automata. The *Tsetlin automaton* shown in Figure 3 provides an example that solves a 2-armed bandit arbitrarily near optimally as $N$ approaches infinity.

It is inconvenient to describe algorithms as finite-state automata, so a move was made to describe the internal state of the agent as a probability distribution according to which actions would be chosen. The probabilities of taking different actions would be adjusted according to their previous successes and failures.

An example, which stands among a set of algorithms independently developed in the mathematical psychology literature (Hilgard & Bower, 1975), is the *linear reward-inaction* algorithm. Let $p_i$ be the agent's probability of taking action $i$.

- When action $a_i$ succeeds,

$$
\begin{aligned}
p_i &:= p_i + \alpha(1 - p_i) \\
p_j &:= p_j - \alpha p_j \text{ for } j \neq i
\end{aligned}
$$

- When action $a_i$ fails, $p_j$ remains unchanged (for all $j$).

This algorithm converges with probability 1 to a vector containing a single 1 and the rest 0's (choosing a particular action with probability 1). Unfortunately, it does not always converge to the correct action; but the probability that it converges to the wrong one can be made arbitrarily small by making $\alpha$ small (Narendra & Thathachar, 1974). There is no literature on the regret of this algorithm.

## 2.2 Ad-Hoc Techniques

In reinforcement-learning practice, some simple, *ad hoc* strategies have been popular. They are rarely, if ever, the best choice for the models of optimality we have used, but they may be viewed as reasonable, computationally tractable, heuristics. Thrun (1992) has surveyed a variety of these techniques.

### 2.2.1 Greedy Strategies

The first strategy that comes to mind is to always choose the action with the highest estimated payoff. The flaw is that early unlucky sampling might indicate that the best action's reward is less than the reward obtained from a suboptimal action. The suboptimal action will always be picked, leaving the true optimal action starved of data and its superiority never discovered. An agent must explore to ameliorate this outcome.

A useful heuristic is *optimism in the face of uncertainty* in which actions are selected greedily, but strongly optimistic prior beliefs are put on their payoffs so that strong negative evidence is needed to eliminate an action from consideration. This still has a measurable danger of starving an optimal but unlucky action, but the risk of this can be made arbitrarily small. Techniques like this have been used in several reinforcement learning algorithms including the interval exploration method (Kaelbling, 1993b) (described shortly), the *exploration bonus* in Dyna (Sutton, 1990), *curiosity-driven exploration* (Schmidhuber, 1991a), and the exploration mechanism in prioritized sweeping (Moore & Atkeson, 1993).

### 2.2.2 Randomized Strategies

Another simple exploration strategy is to take the action with the best estimated expected reward by default, but with probability $p$, choose an action at random. Some versions of this strategy start with a large value of $p$ to encourage initial exploration, which is slowly decreased.

An objection to the simple strategy is that when it experiments with a non-greedy action it is no more likely to try a promising alternative than a clearly hopeless alternative. A slightly more sophisticated strategy is *Boltzmann exploration*. In this case, the expected reward for taking action $a$, $ER(a)$ is used to choose an action probabilistically according to the distribution

$$P(a) = \frac{e^{ER(a)/T}}{\sum_{a' \in A} e^{ER(a')/T}} \quad .$$

The *temperature* parameter $T$ can be decreased over time to decrease exploration. This method works well if the best action is well separated from the others, but suffers somewhat when the values of the actions are close. It may also converge unnecessarily slowly unless the temperature schedule is manually tuned with great care.

### 2.2.3 Interval-based Techniques

Exploration is often more efficient when it is based on second-order information about the certainty or variance of the estimated values of actions. Kaelbling's *interval estimation* algorithm (1993b) stores statistics for each action $a_i$: $w_i$ is the number of successes and $n_i$ the number of trials. An action is chosen by computing the upper bound of a $100 \cdot (1 - \alpha)\%$ confidence interval on the success probability of each action and choosing the action with the highest upper bound. Smaller values of the $\alpha$ parameter encourage greater exploration. When payoffs are boolean, the normal approximation to the binomial distribution can be used to construct the confidence interval (though the binomial should be used for small $n$). Other payoff distributions can be handled using their associated statistics or with nonparametric methods. The method works very well in empirical trials. It is also related to a certain class of statistical techniques known as *experiment design* methods (Box & Draper, 1987), which are used for comparing multiple treatments (for example, fertilizers or drugs) to determine which treatment (if any) is best in as small a set of experiments as possible.

## 2.3 More General Problems

When there are multiple states, but reinforcement is still immediate, then any of the above solutions can be replicated, once for each state. However, when generalization is required, these solutions must be integrated with generalization methods (see section 6); this is straightforward for the simple ad-hoc methods, but it is not understood how to maintain theoretical guarantees.

Many of these techniques focus on converging to some regime in which exploratory actions are taken rarely or never; this is appropriate when the environment is stationary. However, when the environment is non-stationary, exploration must continue to take place, in order to notice changes in the world. Again, the more ad-hoc techniques can be modified to deal with this in a plausible manner (keep temperature parameters from going to 0; decay the statistics in interval estimation), but none of the theoretically guaranteed methods can be applied.

## 3. Delayed Reward

In the general case of the reinforcement learning problem, the agent's actions determine not only its immediate reward, but also (at least probabilistically) the next state of the environment. Such environments can be thought of as networks of bandit problems, but the agent must take into account the next state as well as the immediate reward when it decides which action to take. The model of long-run optimality the agent is using determines exactly how it should take the value of the future into account. The agent will have to be able to learn from delayed reinforcement: it may take a long sequence of actions, receiving insignificant reinforcement, then finally arrive at a state with high reinforcement. The agent must be able to learn which of its actions are desirable based on reward that can take place arbitrarily far in the future.

### 3.1 Markov Decision Processes

Problems with delayed reinforcement are well modeled as *Markov decision processes* (MDPs). An MDP consists of

- a set of states $\mathcal{S}$,

- a set of actions $\mathcal{A}$,

- a reward function $R : \mathcal{S} \times \mathcal{A} \to \Re$, and

- a state transition function $T : \mathcal{S} \times \mathcal{A} \to \Pi(\mathcal{S})$, where a member of $\Pi(\mathcal{S})$ is a probability distribution over the set $\mathcal{S}$ (i.e. it maps states to probabilities). We write $T(s, a, s')$ for the probability of making a transition from state $s$ to state $s'$ using action $a$.

The state transition function probabilistically specifies the next state of the environment as a function of its current state and the agent's action. The reward function specifies expected instantaneous reward as a function of the current state and action. The model is *Markov* if the state transitions are independent of any previous environment states or agent actions. There are many good references to MDP models (Bellman, 1957; Bertsekas, 1987; Howard, 1960; Puterman, 1994).

Although general MDPs may have infinite (even uncountable) state and action spaces, we will only discuss methods for solving finite-state and finite-action problems. In section 6, we discuss methods for solving problems with continuous input and output spaces.

## 3.2 Finding a Policy Given a Model

Before we consider algorithms for learning to behave in MDP environments, we will explore techniques for determining the optimal policy given a correct model. These dynamic programming techniques will serve as the foundation and inspiration for the learning algorithms to follow. We restrict our attention mainly to finding optimal policies for the infinite-horizon discounted model, but most of these algorithms have analogs for the finite-horizon and average-case models as well. We rely on the result that, for the infinite-horizon discounted model, there exists an optimal deterministic stationary policy (Bellman, 1957).

We will speak of the optimal *value* of a state—it is the expected infinite discounted sum of reward that the agent will gain if it starts in that state and executes the optimal policy. Using $\pi$ as a complete decision policy, it is written

$$V^*(s) = \max_\pi E \left( \sum_{t=0}^{\infty} \gamma^t r_t \right) \ .$$

This optimal value function is unique and can be defined as the solution to the simultaneous equations

$$V^*(s) = \max_a \left( R(s, a) + \gamma \sum_{s' \in \mathcal{S}} T(s, a, s') V^*(s') \right), \forall s \in \mathcal{S} \ , \tag{1}$$

which assert that the value of a state $s$ is the expected instantaneous reward plus the expected discounted value of the next state, using the best available action. Given the optimal value function, we can specify the optimal policy as

$$\pi^*(s) = \arg\max_a \left( R(s, a) + \gamma \sum_{s' \in \mathcal{S}} T(s, a, s') V^*(s') \right) \ .$$

### 3.2.1 VALUE ITERATION

One way, then, to find an optimal policy is to find the optimal value function. It can be determined by a simple iterative algorithm called *value iteration* that can be shown to converge to the correct $V^*$ values (Bellman, 1957; Bertsekas, 1987).

```
initialize V(s) arbitrarily
loop until policy good enough
    loop for s ∈ S
        loop for a ∈ A
            Q(s,a) := R(s,a) + γ ∑_{s'∈S} T(s,a,s')V(s')
        V(s) := max_a Q(s,a)
    end loop
end loop
```

It is not obvious when to stop the value iteration algorithm. One important result bounds the performance of the current greedy policy as a function of the *Bellman residual* of the current value function (Williams & Baird, 1993b). It says that if the maximum difference between two successive value functions is less than $\epsilon$, then the value of the greedy policy, (the policy obtained by choosing, in every state, the action that maximizes the estimated discounted reward, using the current estimate of the value function) differs from the value function of the optimal policy by no more than $2\epsilon\gamma/(1-\gamma)$ at any state. This provides an effective stopping criterion for the algorithm. Puterman (1994) discusses another stopping criterion, based on the *span semi-norm*, which may result in earlier termination. Another important result is that the greedy policy is guaranteed to be optimal in some finite number of steps even though the value function may not have converged (Bertsekas, 1987). And in practice, the greedy policy is often optimal long before the value function has converged.

Value iteration is very flexible. The assignments to $V$ need not be done in strict order as shown above, but instead can occur asynchronously in parallel provided that the value of every state gets updated infinitely often on an infinite run. These issues are treated extensively by Bertsekas (1989), who also proves convergence results.

Updates based on Equation 1 are known as *full backups* since they make use of information from all possible successor states. It can be shown that updates of the form

$$Q(s,a) := Q(s,a) + \alpha(r + \gamma \max_{a'} Q(s',a') - Q(s,a))$$

can also be used as long as each pairing of $a$ and $s$ is updated infinitely often, $s'$ is sampled from the distribution $T(s,a,s')$, $r$ is sampled with mean $R(s,a)$ and bounded variance, and the learning rate $\alpha$ is decreased slowly. This type of *sample backup* (Singh, 1993) is critical to the operation of the model-free methods discussed in the next section.

The computational complexity of the value-iteration algorithm with full backups, per iteration, is quadratic in the number of states and linear in the number of actions. Commonly, the transition probabilities $T(s,a,s')$ are sparse. If there are on average a constant number of next states with non-zero probability then the cost per iteration is linear in the number of states and linear in the number of actions. The number of iterations required to reach the optimal value function is polynomial in the number of states and the magnitude of the largest reward if the discount factor is held constant. However, in the worst case the number of iterations grows polynomially in $1/(1-\gamma)$, so the convergence rate slows considerably as the discount factor approaches 1 (Littman, Dean, & Kaelbling, 1995b).

### 3.2.2 POLICY ITERATION

The *policy iteration* algorithm manipulates the policy directly, rather than finding it indirectly via the optimal value function. It operates as follows:

```
choose an arbitrary policy π′
loop
     π := π′
     compute the value function of policy π:
          solve the linear equations
               V_π(s) = R(s, π(s)) + γ ∑_{s′∈S} T(s, π(s), s′)V_π(s′)
     improve the policy at each state:
          π′(s) := arg max_a (R(s, a) + γ ∑_{s′∈S} T(s, a, s′)V_π(s′))
until π = π′
```

The value function of a policy is just the expected infinite discounted reward that will be gained, at each state, by executing that policy. It can be determined by solving a set of linear equations. Once we know the value of each state under the current policy, we consider whether the value could be improved by changing the first action taken. If it can, we change the policy to take the new action whenever it is in that situation. This step is guaranteed to strictly improve the performance of the policy. When no improvements are possible, then the policy is guaranteed to be optimal.

Since there are at most $|\mathcal{A}|^{|\mathcal{S}|}$ distinct policies, and the sequence of policies improves at each step, this algorithm terminates in at most an exponential number of iterations (Puterman, 1994). However, it is an important open question how many iterations policy iteration takes in the worst case. It is known that the running time is pseudopolynomial and that for any fixed discount factor, there is a polynomial bound in the total size of the MDP (Littman et al., 1995b).

### 3.2.3 ENHANCEMENT TO VALUE ITERATION AND POLICY ITERATION

In practice, value iteration is much faster per iteration, but policy iteration takes fewer iterations. Arguments have been put forth to the effect that each approach is better for large problems. Puterman's *modified policy iteration* algorithm (Puterman & Shin, 1978) provides a method for trading iteration time for iteration improvement in a smoother way. The basic idea is that the expensive part of policy iteration is solving for the exact value of $V_\pi$. Instead of finding an exact value for $V_\pi$, we can perform a few steps of a modified value-iteration step where the policy is held fixed over successive iterations. This can be shown to produce an approximation to $V_\pi$ that converges linearly in $\gamma$. In practice, this can result in substantial speedups.

Several standard numerical-analysis techniques that speed the convergence of dynamic programming can be used to accelerate value and policy iteration. *Multigrid methods* can be used to quickly seed a good initial approximation to a high resolution value function by initially performing value iteration at a coarser resolution (Rüde, 1993). *State aggregation* works by collapsing groups of states to a single meta-state solving the abstracted problem (Bertsekas & Castañon, 1989).

### 3.2.4 COMPUTATIONAL COMPLEXITY

Value iteration works by producing successive approximations of the optimal value function. Each iteration can be performed in $O(|A||S|^2)$ steps, or faster if there is sparsity in the transition function. However, the number of iterations required can grow exponentially in the discount factor (Condon, 1992); as the discount factor approaches 1, the decisions must be based on results that happen farther and farther into the future. In practice, policy iteration converges in fewer iterations than value iteration, although the per-iteration costs of $O(|A||S|^2 + |S|^3)$ can be prohibitive. There is no known tight worst-case bound available for policy iteration (Littman et al., 1995b). Modified policy iteration (Puterman & Shin, 1978) seeks a trade-off between cheap and effective iterations and is preferred by some practictioners (Rust, 1996).

Linear programming (Schrijver, 1986) is an extremely general problem, and MDPs can be solved by general-purpose linear-programming packages (Derman, 1970; D'Epenoux, 1963; Hoffman & Karp, 1966). An advantage of this approach is that commercial-quality linear-programming packages are available, although the time and space requirements can still be quite high. From a theoretic perspective, linear programming is the only known algorithm that can solve MDPs in polynomial time, although the theoretically efficient algorithms have not been shown to be efficient in practice.

## 4. Learning an Optimal Policy: Model-free Methods

In the previous section we reviewed methods for obtaining an optimal policy for an MDP assuming that we already had a model. The model consists of knowledge of the state transition probability function $T(s, a, s')$ and the reinforcement function $R(s, a)$. Reinforcement learning is primarily concerned with how to obtain the optimal policy when such a model is not known in advance. The agent must interact with its environment directly to obtain information which, by means of an appropriate algorithm, can be processed to produce an optimal policy.

At this point, there are two ways to proceed.

- **Model-free:** Learn a controller without learning a model.

- **Model-based:** Learn a model, and use it to derive a controller.

Which approach is better? This is a matter of some debate in the reinforcement-learning community. A number of algorithms have been proposed on both sides. This question also appears in other fields, such as adaptive control, where the dichotomy is between *direct* and *indirect* adaptive control.

This section examines model-free learning, and Section 5 examines model-based methods.

The biggest problem facing a reinforcement-learning agent is *temporal credit assignment*. How do we know whether the action just taken is a good one, when it might have far-reaching effects? One strategy is to wait until the "end" and reward the actions taken if the result was good and punish them if the result was bad. In ongoing tasks, it is difficult to know what the "end" is, and this might require a great deal of memory. Instead, we will use insights from value iteration to adjust the estimated value of a state based on the immediate reward and the estimated value of the next state. This class of algorithms is known as *temporal difference methods* (Sutton, 1988). We will consider two different temporal-difference learning strategies for the discounted infinite-horizon model.

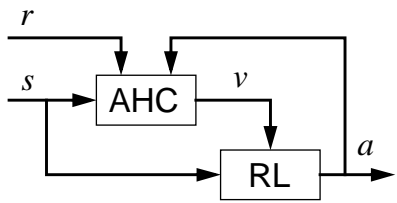

Figure 4: Architecture for the adaptive heuristic critic.

## 4.1 Adaptive Heuristic Critic and TD($\lambda$)

The *adaptive heuristic critic* algorithm is an adaptive version of policy iteration (Barto, Sutton, & Anderson, 1983) in which the value-function computation is no longer implemented by solving a set of linear equations, but is instead computed by an algorithm called $TD(0)$. A block diagram for this approach is given in Figure 4. It consists of two components: a critic (labeled AHC), and a reinforcement-learning component (labeled RL). The reinforcement-learning component can be an instance of any of the $k$-armed bandit algorithms, modified to deal with multiple states and non-stationary rewards. But instead of acting to maximize instantaneous reward, it will be acting to maximize the heuristic value, $v$, that is computed by the critic. The critic uses the real external reinforcement signal to learn to map states to their expected discounted values given that the policy being executed is the one currently instantiated in the RL component.

We can see the analogy with modified policy iteration if we imagine these components working in alternation. The policy $\pi$ implemented by RL is fixed and the critic learns the value function $V_\pi$ for that policy. Now we fix the critic and let the RL component learn a new policy $\pi'$ that maximizes the new value function, and so on. In most implementations, however, both components operate simultaneously. Only the alternating implementation can be guaranteed to converge to the optimal policy, under appropriate conditions. Williams and Baird explored the convergence properties of a class of AHC-related algorithms they call "incremental variants of policy iteration" (Williams & Baird, 1993a).

It remains to explain how the critic can learn the value of a policy. We define $\langle s, a, r, s' \rangle$ to be an *experience tuple* summarizing a single transition in the environment. Here $s$ is the agent's state before the transition, $a$ is its choice of action, $r$ the instantaneous reward it receives, and $s'$ its resulting state. The value of a policy is learned using Sutton's $TD(0)$ algorithm (Sutton, 1988) which uses the update rule

$$V(s) := V(s) + \alpha(r + \gamma V(s') - V(s)) \ .$$

Whenever a state $s$ is visited, its estimated value is updated to be closer to $r + \gamma V(s')$, since $r$ is the instantaneous reward received and $V(s')$ is the estimated value of the actually occurring next state. This is analogous to the sample-backup rule from value iteration—the only difference is that the sample is drawn from the real world rather than by simulating a known model. The key idea is that $r + \gamma V(s')$ is a sample of the value of $V(s)$, and it is more likely to be correct because it incorporates the real $r$. If the learning rate $\alpha$ is adjusted properly (it must be slowly decreased) and the policy is held fixed, $TD(0)$ is guaranteed to converge to the optimal value function.

The $TD(0)$ rule as presented above is really an instance of a more general class of algorithms called $TD(\lambda)$, with $\lambda = 0$. $TD(0)$ looks only one step ahead when adjusting value estimates; although it will eventually arrive at the correct answer, it can take quite a while to do so. The general $TD(\lambda)$ rule is similar to the $TD(0)$ rule given above,

$$V(u) := V(u) + \alpha(r + \gamma V(s') - V(s))e(u) \ ,$$

but it is applied to *every state* according to its eligibility $e(u)$, rather than just to the immediately previous state, $s$. One version of the eligibility trace is defined to be

$$e(s) = \sum_{k=1}^{t} (\lambda\gamma)^{t-k} \delta_{s,s_k} \ , \text{ where } \delta_{s,s_k} = \left\{ \begin{array}{l} 1 \text{ if } s = s_k \\ 0 \text{ otherwise} \end{array} \right. \ .$$

The eligibility of a state $s$ is the degree to which it has been visited in the recent past; when a reinforcement is received, it is used to update all the states that have been recently visited, according to their eligibility. When $\lambda = 0$ this is equivalent to $TD(0)$. When $\lambda = 1$, it is roughly equivalent to updating all the states according to the number of times they were visited by the end of a run. Note that we can update the eligibility online as follows:

$$e(s) := \left\{ \begin{array}{ll} \gamma\lambda e(s) + 1 & \text{if } s = \text{ current state} \\ \gamma\lambda e(s) & \text{otherwise} \end{array} \right. \ .$$

It is computationally more expensive to execute the general $TD(\lambda)$, though it often converges considerably faster for large $\lambda$ (Dayan, 1992; Dayan & Sejnowski, 1994). There has been some recent work on making the updates more efficient (Cichosz & Mulawka, 1995) and on changing the definition to make $TD(\lambda)$ more consistent with the certainty-equivalent method (Singh & Sutton, 1996), which is discussed in Section 5.1.

## 4.2 Q-learning

The work of the two components of AHC can be accomplished in a unified manner by Watkins' Q-learning algorithm (Watkins, 1989; Watkins & Dayan, 1992). Q-learning is typically easier to implement. In order to understand Q-learning, we have to develop some additional notation. Let $Q^*(s, a)$ be the expected discounted reinforcement of taking action $a$ in state $s$, then continuing by choosing actions optimally. Note that $V^*(s)$ is the value of $s$ assuming the best action is taken initially, and so $V^*(s) = \max_a Q^*(s, a)$. $Q^*(s, a)$ can hence be written recursively as

$$Q^*(s, a) = R(s, a) + \gamma \sum_{s' \in \mathcal{S}} T(s, a, s') \max_{a'} Q^*(s', a') \ .$$

Note also that, since $V^*(s) = \max_a Q^*(s, a)$, we have $\pi^*(s) = \arg\max_a Q^*(s, a)$ as an optimal policy.

Because the $Q$ function makes the action explicit, we can estimate the $Q$ values online using a method essentially the same as $TD(0)$, but also use them to define the policy, because an action can be chosen just by taking the one with the maximum $Q$ value for the current state.

The Q-learning rule is

$$Q(s,a) := Q(s,a) + \alpha(r + \gamma \max_{a'} Q(s',a') - Q(s,a)) \ ,$$

where $\langle s, a, r, s' \rangle$ is an experience tuple as described earlier. If each action is executed in each state an infinite number of times on an infinite run and $\alpha$ is decayed appropriately, the $Q$ values will converge with probability 1 to $Q^*$ (Watkins, 1989; Tsitsiklis, 1994; Jaakkola, Jordan, & Singh, 1994). Q-learning can also be extended to update states that occurred more than one step previously, as in $TD(\lambda)$ (Peng & Williams, 1994).

When the $Q$ values are nearly converged to their optimal values, it is appropriate for the agent to act greedily, taking, in each situation, the action with the highest $Q$ value. During learning, however, there is a difficult exploitation versus exploration trade-off to be made. There are no good, formally justified approaches to this problem in the general case; standard practice is to adopt one of the *ad hoc* methods discussed in section 2.2.

AHC architectures seem to be more difficult to work with than Q-learning on a practical level. It can be hard to get the relative learning rates right in AHC so that the two components converge together. In addition, Q-learning is *exploration insensitive*: that is, that the Q values will converge to the optimal values, independent of how the agent behaves while the data is being collected (as long as all state-action pairs are tried often enough). This means that, although the exploration-exploitation issue must be addressed in Q-learning, the details of the exploration strategy will not affect the convergence of the learning algorithm. For these reasons, Q-learning is the most popular and seems to be the most effective model-free algorithm for learning from delayed reinforcement. It does not, however, address any of the issues involved in generalizing over large state and/or action spaces. In addition, it may converge quite slowly to a good policy.

## 4.3 Model-free Learning With Average Reward

As described, Q-learning can be applied to discounted infinite-horizon MDPs. It can also be applied to undiscounted problems as long as the optimal policy is guaranteed to reach a reward-free absorbing state and the state is periodically reset.

Schwartz (1993) examined the problem of adapting Q-learning to an average-reward framework. Although his R-learning algorithm seems to exhibit convergence problems for some MDPs, several researchers have found the average-reward criterion closer to the true problem they wish to solve than a discounted criterion and therefore prefer R-learning to Q-learning (Mahadevan, 1994).

With that in mind, researchers have studied the problem of learning optimal average-reward policies. Mahadevan (1996) surveyed model-based average-reward algorithms from a reinforcement-learning perspective and found several difficulties with existing algorithms. In particular, he showed that existing reinforcement-learning algorithms for average reward (and some dynamic programming algorithms) do not always produce bias-optimal policies. Jaakkola, Jordan and Singh (1995) described an average-reward learning algorithm with guaranteed convergence properties. It uses a Monte-Carlo component to estimate the expected future reward for each state as the agent moves through the environment. In addition, Bertsekas presents a Q-learning-like algorithm for average-case reward in his new textbook (1995). Although this recent work provides a much needed theoretical foundation to this area of reinforcement learning, many important problems remain unsolved.

## 5. Computing Optimal Policies by Learning Models

The previous section showed how it is possible to learn an optimal policy without knowing the models $T(s, a, s')$ or $R(s, a)$ and without even learning those models en route. Although many of these methods are guaranteed to find optimal policies eventually and use very little computation time per experience, they make extremely inefficient use of the data they gather and therefore often require a great deal of experience to achieve good performance. In this section we still begin by assuming that we don't know the models in advance, but we examine algorithms that do operate by learning these models. These algorithms are especially important in applications in which computation is considered to be cheap and real-world experience costly.

### 5.1 Certainty Equivalent Methods

We begin with the most conceptually straightforward method: first, learn the $T$ and $R$ functions by exploring the environment and keeping statistics about the results of each action; next, compute an optimal policy using one of the methods of Section 3. This method is known as *certainty equivlance* (Kumar & Varaiya, 1986).

There are some serious objections to this method:

- It makes an arbitrary division between the learning phase and the acting phase.

- How should it gather data about the environment initially? Random exploration might be dangerous, and in some environments is an immensely inefficient method of gathering data, requiring exponentially more data (Whitehead, 1991) than a system that interleaves experience gathering with policy-building more tightly (Koenig & Simmons, 1993). See Figure 5 for an example.

- The possibility of changes in the environment is also problematic. Breaking up an agent's life into a pure learning and a pure acting phase has a considerable risk that the optimal controller based on early life becomes, without detection, a suboptimal controller if the environment changes.

A variation on this idea is *certainty equivalence*, in which the model is learned continually through the agent's lifetime and, at each step, the current model is used to compute an optimal policy and value function. This method makes very effective use of available data, but still ignores the question of exploration and is extremely computationally demanding, even for fairly small state spaces. Fortunately, there are a number of other model-based algorithms that are more practical.

### 5.2 Dyna

Sutton's Dyna architecture (1990, 1991) exploits a middle ground, yielding strategies that are both more effective than model-free learning and more computationally efficient than the certainty-equivalence approach. It simultaneously uses experience to build a model ($\hat{T}$ and $\hat{R}$), uses experience to adjust the policy, and uses the model to adjust the policy.

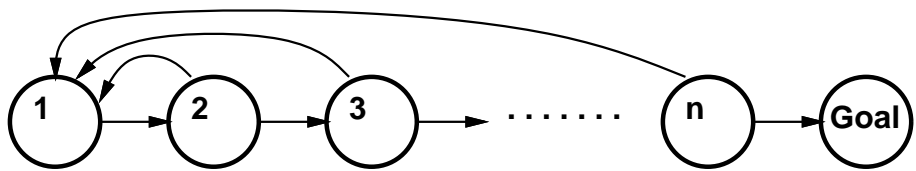

Figure 5: In this environment, due to Whitehead (1991), random exploration would take take $O(2^n)$ steps to reach the goal even once, whereas a more intelligent exploration strategy (e.g. "assume any untried action leads directly to goal") would require only $O(n^2)$ steps.

Dyna operates in a loop of interaction with the environment. Given an experience tuple $\langle s, a, s', r \rangle$, it behaves as follows:

- Update the model, incrementing statistics for the transition from $s$ to $s'$ on action $a$ and for receiving reward $r$ for taking action $a$ in state $s$. The updated models are $\hat{T}$ and $\hat{R}$.

- Update the policy at state $s$ based on the newly updated model using the rule

$$Q(s, a) := \hat{R}(s, a) + \gamma \sum_{s'} \hat{T}(s, a, s') \max_{a'} Q(s', a') \ \ ,$$

which is a version of the value-iteration update for $Q$ values.

- Perform $k$ additional updates: choose $k$ state-action pairs at random and update them according to the same rule as before:

$$Q(s_k, a_k) := \hat{R}(s_k, a_k) + \gamma \sum_{s'} \hat{T}(s_k, a_k, s') \max_{a'} Q(s', a') \ \ .$$

- Choose an action $a'$ to perform in state $s'$, based on the $Q$ values but perhaps modified by an exploration strategy.

The Dyna algorithm requires about $k$ times the computation of Q-learning per instance, but this is typically vastly less than for the naive model-based method. A reasonable value of $k$ can be determined based on the relative speeds of computation and of taking action.

Figure 6 shows a grid world in which in each cell the agent has four actions (N, S, E, W) and transitions are made deterministically to an adjacent cell, unless there is a block, in which case no movement occurs. As we will see in Table 1, Dyna requires an order of magnitude fewer steps of experience than does Q-learning to arrive at an optimal policy. Dyna requires about six times more computational effort, however.

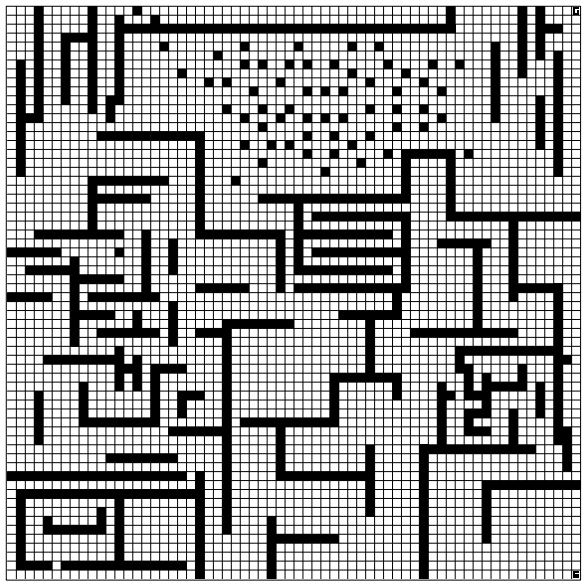

Figure 6: A 3277-state grid world. This was formulated as a shortest-path reinforcement-learning problem, which yields the same result as if a reward of 1 is given at the goal, a reward of zero elsewhere and a discount factor is used.

| | Steps before convergence | Backups before convergence |
|---|---|---|
| Q-learning | 531,000 | 531,000 |
| Dyna | 62,000 | 3,055,000 |
| prioritized sweeping | 28,000 | 1,010,000 |

Table 1: The performance of three algorithms described in the text. All methods used the exploration heuristic of "optimism in the face of uncertainty": any state not previously visited was assumed by default to be a goal state. Q-learning used its optimal learning rate parameter for a deterministic maze: $\alpha = 1$. Dyna and prioritized sweeping were permitted to take $k = 200$ backups per transition. For prioritized sweeping, the priority queue often emptied before all backups were used.

### 5.3 Prioritized Sweeping / Queue-Dyna

Although Dyna is a great improvement on previous methods, it suffers from being relatively undirected. It is particularly unhelpful when the goal has just been reached or when the agent is stuck in a dead end; it continues to update random state-action pairs, rather than concentrating on the "interesting" parts of the state space. These problems are addressed by prioritized sweeping (Moore & Atkeson, 1993) and Queue-Dyna (Peng & Williams, 1993), which are two independently-developed but very similar techniques. We will describe prioritized sweeping in some detail.

The algorithm is similar to Dyna, except that updates are no longer chosen at random and values are now associated with states (as in value iteration) instead of state-action pairs (as in Q-learning). To make appropriate choices, we must store additional information in the model. Each state remembers its *predecessors*: the states that have a non-zero transition probability to it under some action. In addition, each state has a *priority*, initially set to zero.

Instead of updating $k$ random state-action pairs, prioritized sweeping updates $k$ states with the highest priority. For each high-priority state $s$, it works as follows:

- Remember the current value of the state: $V_{old} = V(s)$.

- Update the state's value

$$V(s) := \max_a \left( \hat{R}(s,a) + \gamma \sum_{s'} \hat{T}(s,a,s')V(s') \right) \quad .$$

- Set the state's priority back to 0.

- Compute the value change $\Delta = |V_{old} - V(s)|$.

- Use $\Delta$ to modify the priorities of the predecessors of $s$.

If we have updated the $V$ value for state $s'$ and it has changed by amount $\Delta$, then the immediate predecessors of $s'$ are informed of this event. Any state $s$ for which there exists an action $a$ such that $\hat{T}(s,a,s') \neq 0$ has its priority promoted to $\Delta \cdot \hat{T}(s,a,s')$, unless its priority already exceeded that value.

The global behavior of this algorithm is that when a real-world transition is "surprising" (the agent happens upon a goal state, for instance), then lots of computation is directed to propagate this new information back to relevant predecessor states. When the real-world transition is "boring" (the actual result is very similar to the predicted result), then computation continues in the most deserving part of the space.

Running prioritized sweeping on the problem in Figure 6, we see a large improvement over Dyna. The optimal policy is reached in about half the number of steps of experience and one-third the computation as Dyna required (and therefore about 20 times fewer steps and twice the computational effort of Q-learning).

## 5.4 Other Model-Based Methods

Methods proposed for solving MDPs given a model can be used in the context of model-based methods as well.

RTDP (real-time dynamic programming) (Barto, Bradtke, & Singh, 1995) is another model-based method that uses Q-learning to concentrate computational effort on the areas of the state-space that the agent is most likely to occupy. It is specific to problems in which the agent is trying to achieve a particular goal state and the reward everywhere else is 0. By taking into account the start state, it can find a short path from the start to the goal, without necessarily visiting the rest of the state space.

The Plexus planning system (Dean, Kaelbling, Kirman, & Nicholson, 1993; Kirman, 1994) exploits a similar intuition. It starts by making an approximate version of the MDP which is much smaller than the original one. The approximate MDP contains a set of states, called the *envelope*, that includes the agent's current state and the goal state, if there is one. States that are not in the envelope are summarized by a single "out" state. The planning process is an alternation between finding an optimal policy on the approximate MDP and adding useful states to the envelope. Action may take place in parallel with planning, in which case irrelevant states are also pruned out of the envelope.

## 6. Generalization

All of the previous discussion has tacitly assumed that it is possible to enumerate the state and action spaces and store tables of values over them. Except in very small environments, this means impractical memory requirements. It also makes inefficient use of experience. In a large, smooth state space we generally expect similar states to have similar values and similar optimal actions. Surely, therefore, there should be some more compact representation than a table. Most problems will have continuous or large discrete state spaces; some will have large or continuous action spaces. The problem of learning in large spaces is addressed through *generalization techniques*, which allow compact storage of learned information and transfer of knowledge between "similar" states and actions.

The large literature of generalization techniques from inductive concept learning can be applied to reinforcement learning. However, techniques often need to be tailored to specific details of the problem. In the following sections, we explore the application of standard function-approximation techniques, adaptive resolution models, and hierarchical methods to the problem of reinforcement learning.

The reinforcement-learning architectures and algorithms discussed above have included the storage of a variety of mappings, including $\mathcal{S} \to \mathcal{A}$ (policies), $\mathcal{S} \to \Re$ (value functions), $\mathcal{S} \times \mathcal{A} \to \Re$ (Q functions and rewards), $\mathcal{S} \times \mathcal{A} \to \mathcal{S}$ (deterministic transitions), and $\mathcal{S} \times \mathcal{A} \times \mathcal{S} \to [0, 1]$ (transition probabilities). Some of these mappings, such as transitions and immediate rewards, can be learned using straightforward supervised learning, and can be handled using any of the wide variety of function-approximation techniques for supervised learning that support noisy training examples. Popular techniques include various neural-network methods (Rumelhart & McClelland, 1986), fuzzy logic (Berenji, 1991; Lee, 1991). CMAC (Albus, 1981), and local memory-based methods (Moore, Atkeson, & Schaal, 1995), such as generalizations of nearest neighbor methods. Other mappings, especially the policy mapping, typically need specialized algorithms because training sets of input-output pairs are not available.

## 6.1 Generalization over Input

A reinforcement-learning agent's current state plays a central role in its selection of reward-maximizing actions. Viewing the agent as a state-free black box, a description of the current state is its input. Depending on the agent architecture, its output is either an action selection, or an evaluation of the current state that can be used to select an action. The problem of deciding how the different aspects of an input affect the value of the output is sometimes called the "structural credit-assignment" problem. This section examines approaches to generating actions or evaluations as a function of a description of the agent's current state.

The first group of techniques covered here is specialized to the case when reward is not delayed; the second group is more generally applicable.

### 6.1.1 IMMEDIATE REWARD

When the agent's actions do not influence state transitions, the resulting problem becomes one of choosing actions to maximize immediate reward as a function of the agent's current state. These problems bear a resemblance to the bandit problems discussed in Section 2 except that the agent should condition its action selection on the current state. For this reason, this class of problems has been described as *associative* reinforcement learning.

The algorithms in this section address the problem of learning from immediate boolean reinforcement where the state is vector valued and the action is a boolean vector. Such algorithms can and have been used in the context of a delayed reinforcement, for instance, as the RL component in the AHC architecture described in Section 4.1. They can also be generalized to real-valued reward through *reward comparison* methods (Sutton, 1984).

**CRBP**  The complementary reinforcement backpropagation algorithm (Ackley & Littman, 1990) (CRBP) consists of a feed-forward network mapping an encoding of the state to an encoding of the action. The action is determined probabilistically from the activation of the output units: if output unit $i$ has activation $y_i$, then bit $i$ of the action vector has value 1 with probability $y_i$, and 0 otherwise. Any neural-network supervised training procedure can be used to adapt the network as follows. If the result of generating action $a$ is $r = 1$, then the network is trained with input-output pair $\langle s, a \rangle$. If the result is $r = 0$, then the network is trained with input-output pair $\langle s, \bar{a} \rangle$, where $\bar{a} = (1 - a_1, \ldots, 1 - a_n)$.

The idea behind this training rule is that whenever an action fails to generate reward, CRBP will try to generate an action that is different from the current choice. Although it seems like the algorithm might oscillate between an action and its complement, that does not happen. One step of training a network will only change the action slightly and since the output probabilities will tend to move toward 0.5, this makes action selection more random and increases search. The hope is that the random distribution will generate an action that works better, and then that action will be reinforced.

**ARC**  The associative reinforcement comparison (ARC) algorithm (Sutton, 1984) is an instance of the AHC architecture for the case of boolean actions, consisting of two feed-forward networks. One learns the value of situations, the other learns a policy. These can be simple linear networks or can have hidden units.

In the simplest case, the entire system learns only to optimize immediate reward. First, let us consider the behavior of the network that learns the policy, a mapping from a vector describing $s$ to a 0 or 1. If the output unit has activation $y_i$, then $a$, the action generated, will be 1 if $y + \nu > 0$, where $\nu$ is normal noise, and 0 otherwise.

The adjustment for the output unit is, in the simplest case,

$$e = r(a - 1/2) \ \ ,$$

where the first factor is the reward received for taking the most recent action and the second encodes which action was taken. The actions are encoded as 0 and 1, so $a - 1/2$ always has the same magnitude; if the reward and the action have the same sign, then action 1 will be made more likely, otherwise action 0 will be.

As described, the network will tend to seek actions that given positive reward. To extend this approach to maximize reward, we can compare the reward to some baseline, $b$. This changes the adjustment to

$$e = (r - b)(a - 1/2) \ \ ,$$

where $b$ is the output of the second network. The second network is trained in a standard supervised mode to estimate $r$ as a function of the input state $s$.

Variations of this approach have been used in a variety of applications (Anderson, 1986; Barto et al., 1983; Lin, 1993b; Sutton, 1984).

**REINFORCE Algorithms**   Williams (1987, 1992) studied the problem of choosing actions to maximize immediate reward. He identified a broad class of update rules that perform gradient descent on the expected reward and showed how to integrate these rules with backpropagation. This class, called REINFORCE algorithms, includes linear reward-inaction (Section 2.1.3) as a special case.

The generic REINFORCE update for a parameter $w_{ij}$ can be written

$$\Delta w_{ij} = \alpha_{ij}(r - b_{ij})\frac{\partial}{\partial w_{ij}}\ln(g_j)$$

where $\alpha_{ij}$ is a non-negative factor, $r$ the current reinforcement, $b_{ij}$ a reinforcement baseline, and $g_i$ is the probability density function used to randomly generate actions based on unit activations. Both $\alpha_{ij}$ and $b_{ij}$ can take on different values for each $w_{ij}$, however, when $\alpha_{ij}$ is constant throughout the system, the expected update is exactly in the direction of the expected reward gradient. Otherwise, the update is in the same half space as the gradient but not necessarily in the direction of steepest increase.

Williams points out that the choice of baseline, $b_{ij}$, can have a profound effect on the convergence speed of the algorithm.

**Logic-Based Methods**   Another strategy for generalization in reinforcement learning is to reduce the learning problem to an associative problem of learning boolean functions. A boolean function has a vector of boolean inputs and a single boolean output. Taking inspiration from mainstream machine learning work, Kaelbling developed two algorithms for learning boolean functions from reinforcement: one uses the bias of $k$-DNF to drive the generalization process (Kaelbling, 1994b); the other searches the space of syntactic descriptions of functions using a simple generate-and-test method (Kaelbling, 1994a).

The restriction to a single boolean output makes these techniques difficult to apply. In very benign learning situations, it is possible to extend this approach to use a collection of learners to independently learn the individual bits that make up a complex output. In general, however, that approach suffers from the problem of very unreliable reinforcement: if a single learner generates an inappropriate output bit, all of the learners receive a low reinforcement value. The CASCADE method (Kaelbling, 1993b) allows a collection of learners to be trained collectively to generate appropriate joint outputs; it is considerably more reliable, but can require additional computational effort.

### 6.1.2 DELAYED REWARD

Another method to allow reinforcement-learning techniques to be applied in large state spaces is modeled on value iteration and Q-learning. Here, a function approximator is used to represent the value function by mapping a state description to a value.

Many reseachers have experimented with this approach: Boyan and Moore (1995) used local memory-based methods in conjunction with value iteration; Lin (1991) used backpropagation networks for Q-learning; Watkins (1989) used CMAC for Q-learning; Tesauro (1992, 1995) used backpropagation for learning the value function in backgammon (described in Section 8.1); Zhang and Dietterich (1995) used backpropagation and $TD(\lambda)$ to learn good strategies for job-shop scheduling.

Although there have been some positive examples, in general there are unfortunate interactions between function approximation and the learning rules. In discrete environments there is a guarantee that any operation that updates the value function (according to the Bellman equations) can only reduce the error between the current value function and the optimal value function. This guarantee no longer holds when generalization is used. These issues are discussed by Boyan and Moore (1995), who give some simple examples of value function errors growing arbitrarily large when generalization is used with value iteration. Their solution to this, applicable only to certain classes of problems, discourages such divergence by only permitting updates whose estimated values can be shown to be near-optimal via a battery of Monte-Carlo experiments.

Thrun and Schwartz (1993) theorize that function approximation of value functions is also dangerous because the errors in value functions due to generalization can become compounded by the "max" operator in the definition of the value function.

Several recent results (Gordon, 1995; Tsitsiklis & Van Roy, 1996) show how the appropriate choice of function approximator can guarantee convergence, though not necessarily to the optimal values. Baird's *residual gradient* technique (Baird, 1995) provides guaranteed convergence to locally optimal solutions.

Perhaps the gloominess of these counter-examples is misplaced. Boyan and Moore (1995) report that their counter-examples *can* be made to work with problem-specific hand-tuning despite the unreliability of untuned algorithms that provably converge in discrete domains. Sutton (1996) shows how modified versions of Boyan and Moore's examples can converge successfully. An open question is whether general principles, ideally supported by theory, can help us understand when value function approximation will succeed. In Sutton's com-parative experiments with Boyan and Moore's counter-examples, he changes four aspects of the experiments:

1. Small changes to the task specifications.

2. A very different kind of function approximator (CMAC (Albus, 1975)) that has weak generalization.

3. A different learning algorithm: SARSA (Rummery & Niranjan, 1994) instead of value iteration.

4. A different training regime. Boyan and Moore sampled states uniformly in state space, whereas Sutton's method sampled along empirical trajectories.

There are intuitive reasons to believe that the fourth factor is particularly important, but more careful research is needed.

**Adaptive Resolution Models**   In many cases, what we would like to do is partition the environment into regions of states that can be considered the same for the purposes of learning and generating actions. Without detailed prior knowledge of the environment, it is very difficult to know what granularity or placement of partitions is appropriate. This problem is overcome in methods that use adaptive resolution; during the course of learning, a partition is constructed that is appropriate to the environment.

**Decision Trees**   In environments that are characterized by a set of boolean or discrete-valued variables, it is possible to learn compact decision trees for representing $Q$ values. The *G-learning* algorithm (Chapman & Kaelbling, 1991), works as follows. It starts by assuming that no partitioning is necessary and tries to learn $Q$ values for the entire environment as if it were one state. In parallel with this process, it gathers statistics based on individual input bits; it asks the question whether there is some bit $b$ in the state description such that the $Q$ values for states in which $b = 1$ are significantly different from $Q$ values for states in which $b = 0$. If such a bit is found, it is used to split the decision tree. Then, the process is repeated in each of the leaves. This method was able to learn very small representations of the $Q$ function in the presence of an overwhelming number of irrelevant, noisy state attributes. It outperformed Q-learning with backpropagation in a simple video-game environment and was used by McCallum (1995) (in conjunction with other techniques for dealing with partial observability) to learn behaviors in a complex driving-simulator. It cannot, however, acquire partitions in which attributes are only significant in combination (such as those needed to solve parity problems).

**Variable Resolution Dynamic Programming**   The VRDP algorithm (Moore, 1991) enables conventional dynamic programming to be performed in real-valued multivariate state-spaces where straightforward discretization would fall prey to the curse of dimensionality. A *kd*-tree (similar to a decision tree) is used to partition state space into coarse regions. The coarse regions are refined into detailed regions, but only in parts of the state space which are predicted to be important. This notion of importance is obtained by running "mental trajectories" through state space. This algorithm proved effective on a number of problems for which full high-resolution arrays would have been impractical. It has the disadvantage of requiring a guess at an initially valid trajectory through state-space.

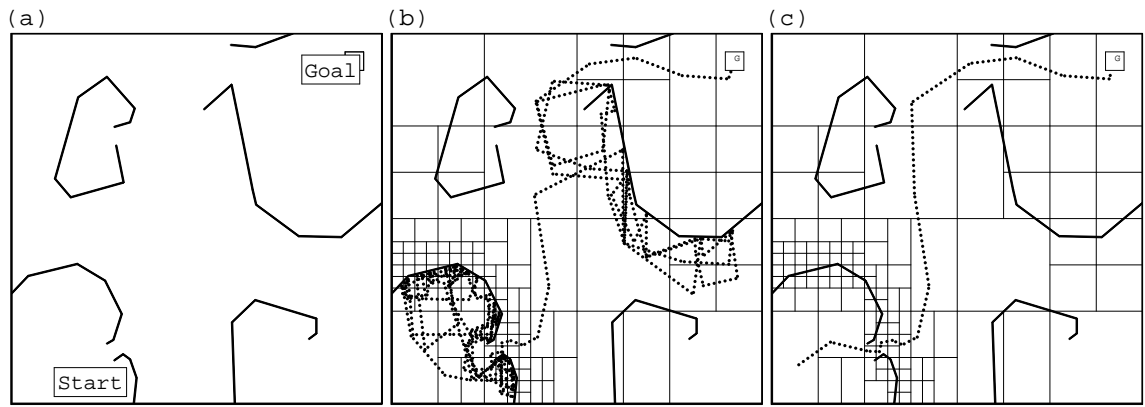

Figure 7: **(a)** A two-dimensional maze problem. The point robot must find a path from start to goal without crossing any of the barrier lines. **(b)** The path taken by PartiGame during the entire first trial. It begins with intense exploration to find a route out of the almost entirely enclosed start region. Having eventually reached a sufficiently high resolution, it discovers the gap and proceeds greedily towards the goal, only to be temporarily blocked by the goal's barrier region. **(c)** The second trial.

**PartiGame Algorithm** Moore's PartiGame algorithm (Moore, 1994) is another solution to the problem of learning to achieve goal configurations in deterministic high-dimensional continuous spaces by learning an adaptive-resolution model. It also divides the environment into cells; but in each cell, the actions available consist of aiming at the neighboring cells (this aiming is accomplished by a local controller, which must be provided as part of the problem statement). The graph of cell transitions is solved for shortest paths in an online incremental manner, but a minimax criterion is used to detect when a group of cells is too coarse to prevent movement between obstacles or to avoid limit cycles. The offending cells are split to higher resolution. Eventually, the environment is divided up just enough to choose appropriate actions for achieving the goal, but no unnecessary distinctions are made. An important feature is that, as well as reducing memory and computational requirements, it also structures exploration of state space in a multi-resolution manner. Given a failure, the agent will initially try something very different to rectify the failure, and only resort to small local changes when all the qualitatively different strategies have been exhausted.

Figure 7a shows a two-dimensional continuous maze. Figure 7b shows the performance of a robot using the PartiGame algorithm during the very first trial. Figure 7c shows the second trial, started from a slightly different position.

This is a very fast algorithm, learning policies in spaces of up to nine dimensions in less than a minute. The restriction of the current implementation to deterministic environments limits its applicability, however. McCallum (1995) suggests some related tree-structured methods.

## 6.2 Generalization over Actions

The networks described in Section 6.1.1 generalize over state descriptions presented as inputs. They also produce outputs in a discrete, factored representation and thus could be seen as generalizing over actions as well.

In cases such as this when actions are described combinatorially, it is important to generalize over actions to avoid keeping separate statistics for the huge number of actions that can be chosen. In continuous action spaces, the need for generalization is even more pronounced.

When estimating $Q$ values using a neural network, it is possible to use either a distinct network for each action, or a network with a distinct output for each action. When the action space is continuous, neither approach is possible. An alternative strategy is to use a single network with both the state and action as input and $Q$ value as the output. Training such a network is not conceptually difficult, but using the network to find the optimal action can be a challenge. One method is to do a local gradient-ascent search on the action in order to find one with high value (Baird & Klopf, 1993).

Gullapalli (1990, 1992) has developed a "neural" reinforcement-learning unit for use in continuous action spaces. The unit generates actions with a normal distribution; it adjusts the mean and variance based on previous experience. When the chosen actions are not performing well, the variance is high, resulting in exploration of the range of choices. When an action performs well, the mean is moved in that direction and the variance decreased, resulting in a tendency to generate more action values near the successful one. This method was successfully employed to learn to control a robot arm with many continuous degrees of freedom.

## 6.3 Hierarchical Methods

Another strategy for dealing with large state spaces is to treat them as a hierarchy of learning problems. In many cases, hierarchical solutions introduce slight sub-optimality in performance, but potentially gain a good deal of efficiency in execution time, learning time, and space.

Hierarchical learners are commonly structured as *gated behaviors*, as shown in Figure 8. There is a collection of *behaviors* that map environment states into low-level actions and a *gating function* that decides, based on the state of the environment, which behavior's actions should be switched through and actually executed. Maes and Brooks (1990) used a version of this architecture in which the individual behaviors were fixed *a priori* and the gating function was learned from reinforcement. Mahadevan and Connell (1991b) used the dual approach: they fixed the gating function, and supplied reinforcement functions for the individual behaviors, which were learned. Lin (1993a) and Dorigo and Colombetti (1995, 1994) both used this approach, first training the behaviors and then training the gating function. Many of the other hierarchical learning methods can be cast in this framework.

### 6.3.1 FEUDAL Q-LEARNING

Feudal Q-learning (Dayan & Hinton, 1993; Watkins, 1989) involves a hierarchy of learning modules. In the simplest case, there is a high-level master and a low-level slave. The master receives reinforcement from the external environment. Its actions consist of commands that it can give to the low-level learner. When the master generates a particular command to the slave, it must reward the slave for taking actions that satisfy the command, even if they do not result in external reinforcement. The master, then, learns a mapping from states to commands. The slave learns a mapping from commands and states to external actions. The set of "commands" and their associated reinforcement functions are established in advance of the learning.

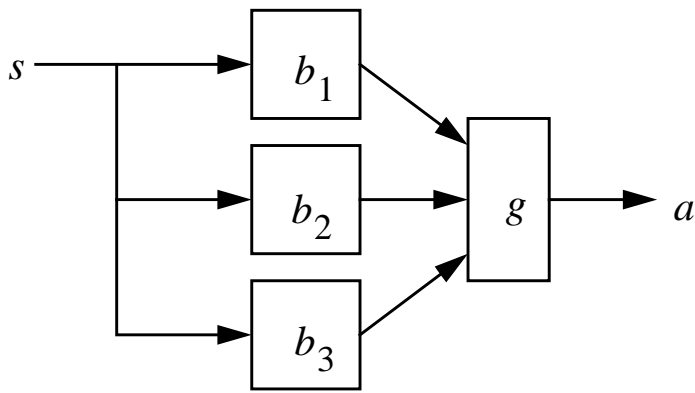

Figure 8: A structure of gated behaviors.

This is really an instance of the general "gated behaviors" approach, in which the slave can execute any of the behaviors depending on its command. The reinforcement functions for the individual behaviors (commands) are given, but learning takes place simultaneously at both the high and low levels.

### 6.3.2 Compositional Q-learning

Singh's compositional Q-learning (1992b, 1992a) (C-QL) consists of a hierarchy based on the temporal sequencing of subgoals. The *elemental tasks* are behaviors that achieve some recognizable condition. The high-level goal of the system is to achieve some set of conditions in sequential order. The achievement of the conditions provides reinforcement for the elemental tasks, which are trained first to achieve individual subgoals. Then, the gating function learns to switch the elemental tasks in order to achieve the appropriate high-level sequential goal. This method was used by Tham and Prager (1994) to learn to control a simulated multi-link robot arm.

### 6.3.3 Hierarchical Distance to Goal

Especially if we consider reinforcement learning modules to be part of larger agent architectures, it is important to consider problems in which goals are dynamically input to the learner. Kaelbling's HDG algorithm (1993a) uses a hierarchical approach to solving problems when goals of achievement (the agent should get to a particular state as quickly as possible) are given to an agent dynamically.

The HDG algorithm works by analogy with navigation in a harbor. The environment is partitioned (*a priori*, but more recent work (Ashar, 1994) addresses the case of learning the partition) into a set of regions whose centers are known as "landmarks." If the agent is currently in the same region as the goal, then it uses low-level actions to move to the goal. If not, then high-level information is used to determine the next landmark on the shortest path from the agent's closest landmark to the goal's closest landmark. Then, the agent uses low-level information to aim toward that next landmark. If errors in action cause deviations in the path, there is no problem; the best aiming point is recomputed on every step.

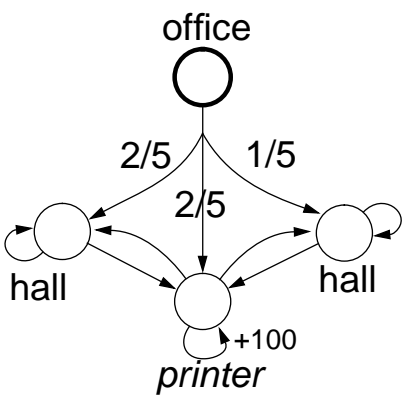

Figure 9: An example of a partially observable environment.

## 7. Partially Observable Environments

In many real-world environments, it will not be possible for the agent to have perfect and complete perception of the state of the environment. Unfortunately, complete observability is necessary for learning methods based on MDPs. In this section, we consider the case in which the agent makes *observations* of the state of the environment, but these observations may be noisy and provide incomplete information. In the case of a robot, for instance, it might observe whether it is in a corridor, an open room, a T-junction, etc., and those observations might be error-prone. This problem is also referred to as the problem of "incomplete perception," "perceptual aliasing," or "hidden state."

In this section, we will consider extensions to the basic MDP framework for solving partially observable problems. The resulting formal model is called a *partially observable Markov decision process* or POMDP.

### 7.1 State-Free Deterministic Policies

The most naive strategy for dealing with partial observability is to ignore it. That is, to treat the observations as if they were the states of the environment and try to learn to behave. Figure 9 shows a simple environment in which the agent is attempting to get to the printer from an office. If it moves from the office, there is a good chance that the agent will end up in one of two places that look like "hall", but that require different actions for getting to the printer. If we consider these states to be the same, then the agent cannot possibly behave optimally. But how well can it do?

The resulting problem is not Markovian, and Q-learning cannot be guaranteed to converge. Small breaches of the Markov requirement are well handled by Q-learning, but it is possible to construct simple environments that cause Q-learning to oscillate (Chrisman &

Littman, 1993). It is possible to use a model-based approach, however; act according to some policy and gather statistics about the transitions between observations, then solve for the optimal policy based on those observations. Unfortunately, when the environment is not Markovian, the transition probabilities depend on the policy being executed, so this new policy will induce a new set of transition probabilities. This approach may yield plausible results in some cases, but again, there are no guarantees.

It is reasonable, though, to ask what the optimal policy (mapping from observations to actions, in this case) is. It is NP-hard (Littman, 1994b) to find this mapping, and even the best mapping can have very poor performance. In the case of our agent trying to get to the printer, for instance, any deterministic state-free policy takes an infinite number of steps to reach the goal on average.

## 7.2 State-Free Stochastic Policies

Some improvement can be gained by considering stochastic policies; these are mappings from observations to probability distributions over actions. If there is randomness in the agent's actions, it will not get stuck in the hall forever. Jaakkola, Singh, and Jordan (1995) have developed an algorithm for finding locally-optimal stochastic policies, but finding a globally optimal policy is still NP hard.

In our example, it turns out that the optimal stochastic policy is for the agent, when in a state that looks like a hall, to go east with probability $2 - \sqrt{2} \approx 0.6$ and west with probability $\sqrt{2} - 1 \approx 0.4$. This policy can be found by solving a simple (in this case) quadratic program. The fact that such a simple example can produce irrational numbers gives some indication that it is a difficult problem to solve exactly.

## 7.3 Policies with Internal State

The only way to behave truly effectively in a wide-range of environments is to use memory of previous actions and observations to disambiguate the current state. There are a variety of approaches to learning policies with internal state.

**Recurrent Q-learning**   One intuitively simple approach is to use a recurrent neural network to learn $Q$ values. The network can be trained using backpropagation through time (or some other suitable technique) and learns to retain "history features" to predict value. This approach has been used by a number of researchers (Meeden, McGraw, & Blank, 1993; Lin & Mitchell, 1992; Schmidhuber, 1991b). It seems to work effectively on simple problems, but can suffer from convergence to local optima on more complex problems.

**Classifier Systems**   Classifier systems (Holland, 1975; Goldberg, 1989) were explicitly developed to solve problems with delayed reward, including those requiring short-term memory. The internal mechanism typically used to pass reward back through chains of decisions, called the *bucket brigade algorithm*, bears a close resemblance to Q-learning. In spite of some early successes, the original design does not appear to handle partially observed environments robustly.

Recently, this approach has been reexamined using insights from the reinforcement-learning literature, with some success. Dorigo did a comparative study of Q-learning and classifier systems (Dorigo & Bersini, 1994). Cliff and Ross (1994) start with Wilson's zeroth-level classifier system (Wilson, 1995) and add one and two-bit memory registers. They find that, although their system can learn to use short-term memory registers effectively, the approach is unlikely to scale to more complex environments.

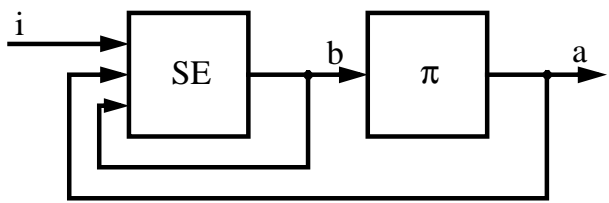

Figure 10: Structure of a POMDP agent.

Dorigo and Colombetti applied classifier systems to a moderately complex problem of learning robot behavior from immediate reinforcement (Dorigo, 1995; Dorigo & Colombetti, 1994).

**Finite-history-window Approach**  One way to restore the Markov property is to allow decisions to be based on the history of recent observations and perhaps actions. Lin and Mitchell (1992) used a fixed-width finite history window to learn a pole balancing task. McCallum (1995) describes the "utile suffix memory" which learns a variable-width window that serves simultaneously as a model of the environment and a finite-memory policy. This system has had excellent results in a very complex driving-simulation domain (McCallum, 1995). Ring (1994) has a neural-network approach that uses a variable history window, adding history when necessary to disambiguate situations.

**POMDP Approach**  Another strategy consists of using hidden Markov model (HMM) techniques to learn a model of the environment, including the hidden state, then to use that model to construct a *perfect memory* controller (Cassandra, Kaelbling, & Littman, 1994; Lovejoy, 1991; Monahan, 1982).

Chrisman (1992) showed how the forward-backward algorithm for learning HMMs could be adapted to learning POMDPs. He, and later McCallum (1993), also gave heuristic *state-splitting rules* to attempt to learn the smallest possible model for a given environment. The resulting model can then be used to integrate information from the agent's observations in order to make decisions.

Figure 10 illustrates the basic structure for a perfect-memory controller. The component on the left is the *state estimator*, which computes the agent's *belief state*, $b$ as a function of the old belief state, the last action $a$, and the current observation $i$. In this context, a belief state is a probability distribution over states of the environment, indicating the likelihood, given the agent's past experience, that the environment is actually in each of those states. The state estimator can be constructed straightforwardly using the estimated world model and Bayes' rule.

Now we are left with the problem of finding a policy mapping belief states into action. This problem can be formulated as an MDP, but it is difficult to solve using the techniques described earlier, because the input space is continuous. Chrisman's approach (1992) does not take into account future uncertainty, but yields a policy after a small amount of computation. A standard approach from the operations-research literature is to solve for the optimal policy (or a close approximation thereof) based on its representation as a piecewise-linear and convex function over the belief space. This method is computationally intractable, but may serve as inspiration for methods that make further approximations (Cassandra et al., 1994; Littman, Cassandra, & Kaelbling, 1995a).

## 8. Reinforcement Learning Applications

One reason that reinforcement learning is popular is that is serves as a theoretical tool for studying the principles of agents learning to act. But it is unsurprising that it has also been used by a number of researchers as a practical computational tool for constructing autonomous systems that improve themselves with experience. These applications have ranged from robotics, to industrial manufacturing, to combinatorial search problems such as computer game playing.

Practical applications provide a test of the efficacy and usefulness of learning algorithms. They are also an inspiration for deciding which components of the reinforcement learning framework are of practical importance. For example, a researcher with a real robotic task can provide a data point to questions such as:

- How important is optimal exploration? Can we break the learning period into exploration phases and exploitation phases?

- What is the most useful model of long-term reward: Finite horizon? Discounted? Infinite horizon?

- How much computation is available between agent decisions and how should it be used?

- What prior knowledge can we build into the system, and which algorithms are capable of using that knowledge?

Let us examine a set of practical applications of reinforcement learning, while bearing these questions in mind.

### 8.1 Game Playing

Game playing has dominated the Artificial Intelligence world as a problem domain ever since the field was born. Two-player games do not fit into the established reinforcement-learning framework since the optimality criterion for games is not one of maximizing reward in the face of a fixed environment, but one of maximizing reward against an optimal adversary (minimax). Nonetheless, reinforcement-learning algorithms can be adapted to work for a very general class of games (Littman, 1994a) and many researchers have used reinforcement learning in these environments. One application, spectacularly far ahead of its time, was Samuel's checkers playing system (Samuel, 1959). This learned a value function represented by a linear function approximator, and employed a training scheme similar to the updates used in value iteration, temporal differences and Q-learning.

More recently, Tesauro (1992, 1994, 1995) applied the temporal difference algorithm to backgammon. Backgammon has approximately $10^{20}$ states, making table-based reinforcement learning impossible. Instead, Tesauro used a backpropagation-based three-layer

|  | Training Games | Hidden Units | Results |
|---|---|---|---|
| Basic |  |  | Poor |
| TD 1.0 | 300,000 | 80 | Lost by 13 points in 51 games |
| TD 2.0 | 800,000 | 40 | Lost by 7 points in 38 games |
| TD 2.1 | 1,500,000 | 80 | Lost by 1 point in 40 games |

Table 2: TD-Gammon's performance in games against the top human professional players. A backgammon tournament involves playing a series of games for points until one player reaches a set target. TD-Gammon won none of these tournaments but came sufficiently close that it is now considered one of the best few players in the world.

neural network as a function approximator for the value function

*Board Position → Probability of victory for current player.*

Two versions of the learning algorithm were used. The first, which we will call Basic TD-Gammon, used very little predefined knowledge of the game, and the representation of a board position was virtually a raw encoding, sufficiently powerful only to permit the neural network to distinguish between conceptually different positions. The second, TD-Gammon, was provided with the same raw state information supplemented by a number of hand-crafted features of backgammon board positions. Providing hand-crafted features in this manner is a good example of how inductive biases from human knowledge of the task can be supplied to a learning algorithm.

The training of both learning algorithms required several months of computer time, and was achieved by constant self-play. No exploration strategy was used—the system always greedily chose the move with the largest expected probability of victory. This naive exploration strategy proved entirely adequate for this environment, which is perhaps surprising given the considerable work in the reinforcement-learning literature which has produced numerous counter-examples to show that greedy exploration can lead to poor learning performance. Backgammon, however, has two important properties. Firstly, whatever policy is followed, every game is guaranteed to end in finite time, meaning that useful reward information is obtained fairly frequently. Secondly, the state transitions are sufficiently stochastic that independent of the policy, all states will occasionally be visited—a wrong initial value function has little danger of starving us from visiting a critical part of state space from which important information could be obtained.

The results (Table 2) of TD-Gammon are impressive. It has competed at the very top level of international human play. Basic TD-Gammon played respectably, but not at a professional standard.

Figure 11: Schaal and Atkeson's devil-sticking robot. The tapered stick is hit alternately by each of the two hand sticks. The task is to keep the devil stick from falling for as many hits as possible. The robot has three motors indicated by torque vectors $\tau_1, \tau_2, \tau_3$.

Although experiments with other games have in some cases produced interesting learning behavior, no success close to that of TD-Gammon has been repeated. Other games that have been studied include Go (Schraudolph, Dayan, & Sejnowski, 1994) and Chess (Thrun, 1995). It is still an open question as to if and how the success of TD-Gammon can be repeated in other domains.

## 8.2 Robotics and Control

In recent years there have been many robotics and control applications that have used reinforcement learning. Here we will concentrate on the following four examples, although many other interesting ongoing robotics investigations are underway.

1. Schaal and Atkeson (1994) constructed a two-armed robot, shown in Figure 11, that learns to juggle a device known as a devil-stick. This is a complex non-linear control task involving a six-dimensional state space and less than 200 msecs per control decision. After about 40 initial attempts the robot learns to keep juggling for hundreds of hits. A typical human learning the task requires an order of magnitude more practice to achieve proficiency at mere tens of hits.

   The juggling robot learned a world model from experience, which was generalized to unvisited states by a function approximation scheme known as locally weighted regression (Cleveland & Delvin, 1988; Moore & Atkeson, 1992). Between each trial, a form of dynamic programming specific to linear control policies and locally linear transitions was used to improve the policy. The form of dynamic programming is known as linear-quadratic-regulator design (Sage & White, 1977).

2. Mahadevan and Connell (1991a) discuss a task in which a mobile robot pushes large boxes for extended periods of time. Box-pushing is a well-known difficult robotics problem, characterized by immense uncertainty in the results of actions. Q-learning was used in conjunction with some novel clustering techniques designed to enable a higher-dimensional input than a tabular approach would have permitted. The robot learned to perform competitively with the performance of a human-programmed solution. Another aspect of this work, mentioned in Section 6.3, was a pre-programmed breakdown of the monolithic task description into a set of lower level tasks to be learned.

3. Mataric (1994) describes a robotics experiment with, from the viewpoint of theoretical reinforcement learning, an unthinkably high dimensional state space, containing many dozens of degrees of freedom. Four mobile robots traveled within an enclosure collecting small disks and transporting them to a destination region. There were three enhancements to the basic Q-learning algorithm. Firstly, pre-programmed signals called *progress estimators* were used to break the monolithic task into subtasks. This was achieved in a robust manner in which the robots were not forced to use the estimators, but had the freedom to profit from the inductive bias they provided. Secondly, control was decentralized. Each robot learned its own policy independently without explicit communication with the others. Thirdly, state space was brutally quantized into a small number of discrete states according to values of a small number of pre-programmed boolean features of the underlying sensors. The performance of the Q-learned policies were almost as good as a simple hand-crafted controller for the job.

4. Q-learning has been used in an elevator dispatching task (Crites & Barto, 1996). The problem, which has been implemented in simulation only at this stage, involved four elevators servicing ten floors. The objective was to minimize the average squared wait time for passengers, discounted into future time. The problem can be posed as a discrete Markov system, but there are $10^{22}$ states even in the most simplified version of the problem. Crites and Barto used neural networks for function approximation and provided an excellent comparison study of their Q-learning approach against the most popular and the most sophisticated elevator dispatching algorithms. The squared wait time of their controller was approximately 7% less than the best alternative algorithm ("Empty the System" heuristic with a receding horizon controller) and less than half the squared wait time of the controller most frequently used in real elevator systems.

5. The final example concerns an application of reinforcement learning by one of the authors of this survey to a packaging task from a food processing industry. The problem involves filling containers with variable numbers of non-identical products. The product characteristics also vary with time, but can be sensed. Depending on the task, various constraints are placed on the container-filling procedure. Here are three examples:

- The mean weight of all containers produced by a shift must not be below the manufacturer's declared weight $W$.

- The number of containers below the declared weight must be less than $P\%$.
- No containers may be produced below weight $W'$.

Such tasks are controlled by machinery which operates according to various *setpoints*. Conventional practice is that setpoints are chosen by human operators, but this choice is not easy as it is dependent on the current product characteristics and the current task constraints. The dependency is often difficult to model and highly non-linear. The task was posed as a finite-horizon Markov decision task in which the state of the system is a function of the product characteristics, the amount of time remaining in the production shift and the mean wastage and percent below declared in the shift so far. The system was discretized into 200,000 discrete states and local weighted regression was used to learn and generalize a transition model. Prioritized sweeping was used to maintain an optimal value function as each new piece of transition information was obtained. In simulated experiments the savings were considerable, typically with wastage reduced by a factor of ten. Since then the system has been deployed successfully in several factories within the United States.

Some interesting aspects of practical reinforcement learning come to light from these examples. The most striking is that in all cases, to make a real system work it proved necessary to supplement the fundamental algorithm with extra pre-programmed knowledge. Supplying extra knowledge comes at a price: more human effort and insight is required and the system is subsequently less autonomous. But it is also clear that for tasks such as these, a knowledge-free approach would not have achieved worthwhile performance within the finite lifetime of the robots.

What forms did this pre-programmed knowledge take? It included an assumption of linearity for the juggling robot's policy, a manual breaking up of the task into subtasks for the two mobile-robot examples, while the box-pusher also used a clustering technique for the $Q$ values which assumed locally consistent $Q$ values. The four disk-collecting robots additionally used a manually discretized state space. The packaging example had far fewer dimensions and so required correspondingly weaker assumptions, but there, too, the assumption of local piecewise continuity in the transition model enabled massive reductions in the amount of learning data required.

The exploration strategies are interesting too. The juggler used careful statistical analysis to judge where to profitably experiment. However, both mobile robot applications were able to learn well with greedy exploration—always exploiting without deliberate exploration. The packaging task used optimism in the face of uncertainty. None of these strategies mirrors theoretically optimal (but computationally intractable) exploration, and yet all proved adequate.

Finally, it is also worth considering the computational regimes of these experiments. They were all very different, which indicates that the differing computational demands of various reinforcement learning algorithms do indeed have an array of differing applications. The juggler needed to make very fast decisions with low latency between each hit, but had long periods (30 seconds and more) between each trial to consolidate the experiences collected on the previous trial and to perform the more aggressive computation necessary to produce a new reactive controller on the next trial. The box-pushing robot was meant to operate autonomously for hours and so had to make decisions with a uniform length control cycle. The cycle was sufficiently long for quite substantial computations beyond simple Q-learning backups. The four disk-collecting robots were particularly interesting. Each robot had a short life of less than 20 minutes (due to battery constraints) meaning that substantial number crunching was impractical, and any significant combinatorial search would have used a significant fraction of the robot's learning lifetime. The packaging task had easy constraints. One decision was needed every few minutes. This provided opportunities for fully computing the optimal value function for the 200,000-state system between every control cycle, in addition to performing massive cross-validation-based optimization of the transition model being learned.

A great deal of further work is currently in progress on practical implementations of reinforcement learning. The insights and task constraints that they produce will have an important effect on shaping the kind of algorithms that are developed in future.

## 9. Conclusions

There are a variety of reinforcement-learning techniques that work effectively on a variety of small problems. But very few of these techniques scale well to larger problems. This is not because researchers have done a bad job of inventing learning techniques, but because it is very difficult to solve arbitrary problems in the general case. In order to solve highly complex problems, we must give up *tabula rasa* learning techniques and begin to incorporate bias that will give leverage to the learning process.

The necessary bias can come in a variety of forms, including the following:

**shaping:** The technique of shaping is used in training animals (Hilgard & Bower, 1975); a teacher presents very simple problems to solve first, then gradually exposes the learner to more complex problems. Shaping has been used in supervised-learning systems, and can be used to train hierarchical reinforcement-learning systems from the bottom up (Lin, 1991), and to alleviate problems of delayed reinforcement by decreasing the delay until the problem is well understood (Dorigo & Colombetti, 1994; Dorigo, 1995).

**local reinforcement signals:** Whenever possible, agents should be given reinforcement signals that are local. In applications in which it is possible to compute a gradient, rewarding the agent for taking steps up the gradient, rather than just for achieving the final goal, can speed learning significantly (Mataric, 1994).

**imitation:** An agent can learn by "watching" another agent perform the task (Lin, 1991). For real robots, this requires perceptual abilities that are not yet available. But another strategy is to have a human supply appropriate motor commands to a robot through a joystick or steering wheel (Pomerleau, 1993).

**problem decomposition:** Decomposing a huge learning problem into a collection of smaller ones, and providing useful reinforcement signals for the subproblems is a very powerful technique for biasing learning. Most interesting examples of robotic reinforcement learning employ this technique to some extent (Connell & Mahadevan, 1993).

**reflexes:** One thing that keeps agents that know nothing from learning anything is that they have a hard time even finding the interesting parts of the space; they wander around at random never getting near the goal, or they are always "killed" immediately. These problems can be ameliorated by programming a set of "reflexes" that cause the agent to act initially in some way that is reasonable (Mataric, 1994; Singh, Barto, Grupen, & Connolly, 1994). These reflexes can eventually be overridden by more detailed and accurate learned knowledge, but they at least keep the agent alive and pointed in the right direction while it is trying to learn. Recent work by Millan (1996) explores the use of reflexes to make robot learning safer and more efficient.

With appropriate biases, supplied by human programmers or teachers, complex reinforcement-learning problems will eventually be solvable. There is still much work to be done and many interesting questions remaining for learning techniques and especially regarding methods for approximating, decomposing, and incorporating bias into problems.

## Acknowledgements

Thanks to Marco Dorigo and three anonymous reviewers for comments that have helped to improve this paper. Also thanks to our many colleagues in the reinforcement-learning community who have done this work and explained it to us.

Leslie Pack Kaelbling was supported in part by NSF grants IRI-9453383 and IRI-9312395. Michael Littman was supported in part by Bellcore. Andrew Moore was supported in part by an NSF Research Initiation Award and by 3M Corporation.